\documentclass{article}
\usepackage[square,sort,comma,numbers]{natbib}

    \usepackage[preprint]{neurips_2020}

\usepackage[utf8]{inputenc} 
\usepackage[T1]{fontenc}    
\usepackage{hyperref}       
\usepackage{url}            
\usepackage{booktabs}       
\usepackage{amsfonts}       
\usepackage{nicefrac}       
\usepackage{microtype}      
\usepackage{verbatim}
\usepackage{graphicx}
\usepackage{extramarks}
\usepackage{lineno}
\usepackage[subtle]{savetrees}
\usepackage[disable]{todonotes}

\newcommand\nnfootnote[1]{%
  \begin{NoHyper}
  \renewcommand\thefootnote{}\footnote{#1}%
  \addtocounter{footnote}{-1}%
  \end{NoHyper}
}
\title{Effectiveness of the Recent Advances \\ in Capsule Networks}

\author{%
  Nidhin Harilal$^*$ \\
  Department of Computer Science\\
  University of Colorado, Boulder\\
   \texttt{nidhin.harilal@colorado.edu} \\
   \And

  Rohan Patil$^*$ \\
  Department of Computer Science\\
  University of California, San Diego\\
  \texttt{rpatil@ucsd.edu} \\
   \And

}

\begin{document}

\maketitle
\nnfootnote{* These authors contributed equally to this work}
\begin{abstract}
    Convolutional neural networks (CNNs) have revolutionized the field of deep neural networks. However, recent research has shown that CNNs fail to generalize under various conditions and hence the idea of capsules was introduced in 2011, though the real surge of research started from 2017. In this paper, we present an overview of the recent advances in capsule architecture and routing mechanisms. In addition, we find that the relative focus in recent literature is on modifying routing procedure or architecture as a whole but the study of other finer components, specifically, squash function is wanting. Thus, we also present some new insights regarding the effect of squash functions in performance of the capsule networks. Finally, we conclude by discussing and proposing possible opportunities in the field of capsule networks.
    
\end{abstract}
\todo{'Extreme"-- fine in abstract but need to elaborate later. In general ``neural network" doesn't need to be capitalized.}
\section{Introduction}

Over the last few years, neural networks have made remarkable progress in various tasks ranging from vision tasks like image recognition and object segmentation to machine translation tasks. The availability of huge amounts of data has made it possible for Neural Networks to excel in different areas of Computer Vision. The considerable success lies in the fact that CNNs can automatically extract high-level features from images, which are much more powerful than human-designed features.

Capsule Networks (CapsNet) were introduced by Hinton et al.~\cite{hinton2011transforming, sabour2017dynamic}, which addressed the significant limitations of CNNs and showed superior performance on the MNIST~\cite{lecun1998mnist} dataset. Presently, capsule networks are regarded as one of the most promising breakthroughs in deep learning. The primary reason behind this is that capsules' idea provides a much more promising way of dealing with different variations in images, including position, scale, orientation, and lighting, than the currently employed methods in the neural networks community.

Despite capsule networks showing an increased positive impact on various vision tasks~\cite{jaiswal2018capsulegan, deng2018hyperspectral, duarte2018videocapsulenet}, the lack of architectural knowledge has limited researchers to exploit the full potential of this new field. Therefore, this paper aims to provide insights behind the working of capsule networks and critically review the latest advances in this field. First, we briefly discuss the reason behind the introduction of capsule networks, followed by a description of its components. Then, we provide an analysis describing aspects and limitations of various components of different capsule networks, followed by an analysis of augmenting squash functions. Lastly, we conclude our analysis by describing opportunities for further research in this field.  

\section{Convolutional Neural Networks (CNNs)}

Convolutional neural networks (CNN) are feed-forward neural networks that can extract features from data in a hierarchical structure. The architecture of CNN is inspired by visual perception~\cite{Hubel1962ReceptiveFB}. CNN architectures date back decades~\cite{CNN_first}, consisting of convolutional layers that can extract high-level features from images. CNNs detect these features in images and learn how to recognize objects with this information. Layers near the start detect simpler features like edges, and as the layers get deeper, they detect more complex features like eyes, noses, or an entire face in case of face recognition. It then uses all of these features, which it has learned to make a final prediction. Deep CNNs have provided a significant contribution in computer vision tasks such as image classification~\cite{imagenet_CNN, imagenet_CNN2}. They have also been successfully applied to other computer vision fields, such as object detection~\cite{yolo, girshick2014rich, lin2017feature},  face recognition~\cite{face_rec}, etc.

\paragraph{Limitations: }CNNs perform exceptionally great when they are inferenced over images that resemble the dataset~\cite{adv_cnn}. Despite being successful, CNN performs poorly when it receives the same image with a different viewpoint~\cite{inv_cnn, liu2018intriguing}. Convolving kernel across an image ensures invariance, but it doesn't certify equivariance. Max-pooling~\cite{maxpool} was introduced to further aid in creating the positional invariance. On close scrutiny, the pooling operation stacked with a convolutional layer will only detect the features but not preserve any spatial relationships between the detected features. The difference could be explained with a CNN detecting a human face. An average human face will have a pair of eyes, nose, and mouth; however, an image in which these parts are present doesn't qualify it to be an image of a human face. Therefore, a system just detecting certain features in the image without having information about their spatial arrangement could fail in many scenarios. In short, The pooling operation, along with the convolution, was supposed to introduce positional, orientational, and proportional invariances but rather became a cause for them~\cite{liu2018intriguing}. Including more data or using methods like data, augmentation is used to tackle this problem, which ensured that the model was trained on as many viewpoints/ orientations as possible. However, this is a very crude way of handling this.
\todo{If the above example is from a specific pape/talk, please refer. }
\section{Capsule Networks (CapsNet)}
Hinton et al.~\cite{hinton2011transforming} proposed the first capsule networks, which, unlike conventional CNNs, were designed to encode and preserve the underlying spatial information between its learned features. The basic idea behind the introduction of capsules by Hinton et al.~\cite{hinton2011transforming} was to create a neural network capable of performing inverse graphics~\cite{hinton2012does}. Computer graphics deals with generating a visual image from some internal hierarchical representation of geometric data. This internal representation consists of matrices that represent the relative positions and orientation of the geometrical objects. This representation is then converted to an image that is finally rendered on the screen. Hinton and his colleagues in their paper~\cite{hinton2011transforming} argued that humans deconstruct a hierarchical representation of the visual information received through eyes and use this representation for recognition. From a pure machine learning perspective, this means that the network should be able to deconstruct a scene into co-related parts, which can be hierarchically represented~\cite{inv_graphics}. To achieve this, Hinton et al.~\cite{hinton2011transforming} proposed to augment the conventional idea of neural network architecture to reflect the idea of several entities. Each entity encapsulates a certain number of neurons and learns to recognize an implicitly defined visual entity over a limited domain of viewing conditions and deformations. 


\subsection{Evolution of Capsules}

In the first proposed capsule networks~\cite{hinton2011transforming}, the output of each capsule consisted of the probability that a specific feature exists along with a set of \textit{instantiating parameters}\footnote{\label{foot:ins_par}Instantiating parameters~\cite{hinton2011transforming} may include and encode the underlying pose, lighting, and deformation of a particular feature relative to the other features that are detected by the capsules in the image.}. The goal of Hinton et al.~\cite{hinton2011transforming} was not recognizing or classifying the objects in an image but rather to force the outputs of a capsule to recognize the pose in an image. A significant limitation of this first implementation of capsule networks was that it required an additional external set of inputs that specified how the image had been transformed for each of the entities to work. Although these transformations could be learned in principle, the more significant challenge was to devise a way to instruct each of the capsules to discover the underlying hierarchical relationship between the transformations in a complete end-to-end train setting without explicitly using additional inputs.



To overcome this challenge, Sabour et al.~\cite{sabour2017dynamic} proposed a new definition for capsules. Each capsule consisted of a series of convolutional layers followed by a fully-connected layer. Unlike the output from conventional artificial neurons, which are scalar, the output from their proposed capsules is represented as an activity vector. The length of this vector encoded the probability of detecting a feature, and the state of the detected feature, also referred to as instantiating parameters, is encoded as the direction of these vectors. To represent the capsule network as vectors whose length represents probabilities, they used a non-linear "squashing" function described in Equation~\ref{eq:squash}.   

\begin{equation}
\label{eq:squash}
v_j = \frac{{||s_j||}_2^{2}}{1+{||s_j||}_2^{2}} \frac{s_j}{{||s_j||}_2}
\end{equation}
where $v_j$ is the vector output of $j^{th}$ capsule and $s_j$ is its total input.
The above-defined non-linearity ensures that the short vector gets shrunk to zero length and long vectors to slightly below 1. This is the basic definition of capsules, and most further improvements in capsule networks utilized this definition of capsules.  

\subsection{Routing mechanism}
\begin{figure}[!htp]
	\centering
	\includegraphics[width=.8\linewidth]{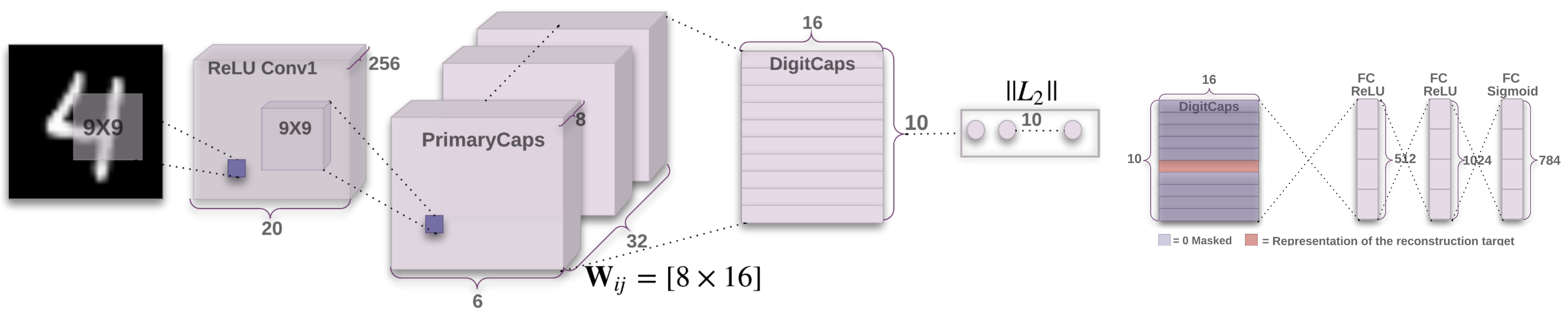}
	\caption{First Routing based Capsule structure (Dynamic Routing) by Sabour et al.~\cite{hinton2011transforming}}
	\label{fig:dyna_cap}
\end{figure}

The routing mechanism is one of the key factors in making the capsule networks work~\cite{sabour2017dynamic}. Routing controls the transfer of information between the capsules. Higher-level capsules get the input from the lower level capsules. The first and the last capsule layers are referred to as the Primary capsule and class capsule layer, respectively. Feature extraction from the image is done by the convolutional layers present inside each capsule, and the output is fed according to the computed squash vectors. Figure~\ref{fig:dyna_cap} first routing method for the capsule network was proposed by Sabour et al.~\cite{sabour2017dynamic}. 

In this, each capsule $i$ (where $1\le i \le N$) in layer $l$ has an activity vector $\mathbf{u_i} \in \mathbb{R}$ to encode the spatial information of features in the form of instantiating parameters using the activation function. This output vector $\mathbf{u_i}$ of the $i^{th}$ lower-level capsule is then fed into all capsules in the neighbouring high-level $(l+1)^{th}$ layer. The total input to a capsule $s_j$ is a weighted sum of "prediction vectors" $\hat{u}_{j|i}$ it received from the lower-level capsules. For all layers of capsules except the primary capsules, The $j^{th}$ capsule at layer $l+1$ will receive $u_i$ and find its product with a corresponding weight matrix $\mathbb{W}_{ij}$ which it is learns during training. 

\begin{equation}
\mathbf{s}_j = \sum_i c_{ij} \hat{u}_{j|i}, \;\; \hat{u}_{j|i} = \mathbb{W}_{ij} \mathbf{u}_{i}      
\end{equation}

Coupling coefficients ($c_{ij}$) ensures that the output of capsule at level $l$ is linked to that in layer $l+1$. These coefficients are determined by a "routing softmax" whose initial logits $b_{ij}$ are the log prior probabilities that capsule $i$ should be coupled to capsule $j$.

\begin{equation}
c_{ij} = \frac{\exp{(b_{ij})}}{\sum_{k} \exp{(b_{ik})}}
\end{equation}

These coefficients are then updated iteratively based on the agreement between the current output $v_j$ of
each capsule, $j$, in the layer above and the prediction $\hat{u}_{j|i}$ made by capsule $i$. In short, lower level capsules are responsible for sending its input to certain higher-level capsules based on the computed coefficients such that it “agrees” in establishing a relation between the input and the task at hand. This is the essence of the dynamic routing method for capsule networks and this mechanism of "agreement" or "voting" between  the capsules is referred as routing-by-agreement~\cite{sabour2017dynamic}.


\section{Advances in Routing Mechanism}
\label{sec:routing}
    The routing method is the distinguishing factor of capsule networks as it provides the mechanism to learn the hierarchial representations, essentially carving out a parse tree. The limitations of dynamic routing for generalizing over complex data have led researchers to introduce other routing methods.
    
    \paragraph{EM Routing:} It is an Expectation-Maximisation routing procedure~\cite{hinton2018matrix}. The two main ideas of this method are: first, the relationship between entity and viewpoint can be encoded as a $4\times4$ matrix, and second, the coefficients will be iteratively updated so that a cluster of similar votes activate the next layer capsules. According to Hinton et al.~\cite{hinton2018matrix}, this method is similar to attention; the clustering is focusing on specific features, but here the direction of attention is reversed in the sense that the competition is between higher layer capsules for votes from lower-level capsules. Another aspect is the robustness of adversarial attacks. Capsule networks with EM Routing show better robustness than CNNs, but interestingly perform similar to a CNN when adversarial examples are generated in a black-box manner using a CNN. 
   \paragraph{KL Routing:} Wang et al.~\cite{wang2018optimization} present a routing procedure as an optimization process with a KL regularization between current and previous state coupling coefficients. For convenience, we call it KL Routing throughout the paper. In their setting, they eliminate the squash function in the iteration part and calculate the squash vector (calculating the squashed vector is different from the original) at the end of iterations. The critical part is that they derive their provided robust mathematical modeling of routing as an agglomerative fuzzy K-Means algorithm~\cite{li2008agglomerative}.
    \paragraph{Self Routing:} It is a Mixture-of-Experts~\cite{nowlan1991evaluation} approach towards routing~\cite{hahn2019self}. The process is non-iterative, thus reducing the computation time. Moreover, it shows better accuracy and robustness towards adversarial attacks as compared to EM routing. The authors also claim that the $4\times4$ matrix multiplication constraint used in EM routing is too restrictive for learning a good representation of complex data with multi-level routing layers. There are two learnable weight matrices $W^{route}$ and $W^{pose}$, which are used for calculating the interaction coefficients and prediction vectors in Self Routing. 
    
\begin{table}[hb]
  \caption{Comparing Routing Procedures.}
  \label{routing_table}
  \centering
  \begin{tabular}{c c c c}
    \toprule
    Routing & Inspiration & Squash function & Iterative \\
    \midrule
    EM Routing & Gaussian Mixture Model & None & Yes \\
    KL Routing & Agglomerative Fuzzy K-Means algorithm & $v_j = \frac{||\sum_i c_{ij} o_{j|i}||}{1 + max_k||\sum_i c_{ik} o_{k|i}||} s_j$ & Yes \\
    Self Routing & Mixture-of-Experts & $v_j = \frac{{||s_j||}_2^{2}}{1+{||s_j||}_2^{2}} \frac{s_j}{{||s_j||}_2}$ & No \\
    \bottomrule
  \end{tabular}\\
  Here {$o_{j|i} = \frac{1}{{||W_{ij}||}_\mathcal{F}} W_{ij} \mu_i$}
\end{table}

\section{Capsule Network Variants and Modifications}
    This section will highlight a few modifications and developments on the original capsule network, which are not limited to routing, summarised in Table \ref{network_table}
    
    \paragraph{CapProNet} In CapProNet, each capsule is associated a subspace of the input vector~\cite{zhang2018cappronet}. Formally, each capsule learns a weight matrix $ W $, the columns of which form the subspace's basis. The input vector is projected on these subspaces before the routing procedure. The authors claim that the advantage of capsule subspaces is that it is possible to find the best direction along which a specific feature's information about the position, orientation, scales, and intrinsic characteristics can be represented better.
    
    \paragraph{RS-CapsNet} The major motivation of RS-CapsNet~\cite{yang2020rs} is from ResNet \cite{he2016deep} and Squeeze-and-Excitation Network \cite{hu2018squeeze}. Their work on routing is majorly to mitigate the issues of stacked capsule layers, which do not perform well~\cite{xi2017capsule}. In RS-CapsNet, a linear combination of the capsule outputs is used for the next layer's input. According to the authors, since the primary capsules are reshaped from convolutional feature maps, redundant background information is present. Taking a linear combination of capsule outputs will reduce the background noise as well as the network will have increased non-linearity.

    \paragraph{Sparse Unsupervised Capsules (SUPCAPS)} The work of Rawlinson et al.~\cite{rawlinson2018sparse} shows how capsules can be used in an unsupervised setting. Furthermore, their work shows the importance of using sparsity in unsupervised learning. In their approach to creating a generator, they initially removed the mask and margin loss function, but the result was that the network started to act as an autoencoder. Furthermore, individual learning capsules were unable to learn about specific features.
    To mitigate this issue, they added dropout as it was observed that the coefficients of capsules were similar, implying that the capsules were unable to carve a parse tree. Adding dropout restored the ability to route in capsules. The discovery that latent capsules lose their ability to specialize in their identity matches the observation that stacked capsule layers have decreased performance \cite{xi2017capsule}.

\begin{table}[!htbp]
  \caption{Capsule Network Varinats and Modifications}
  \label{network_table}
  \centering
  \begin{tabular}{p{0.13\linewidth} p{0.3\linewidth} p{0.3\linewidth}}
    \toprule
    Variants & Modification & Contributions\\
    \midrule
    CapProNet & Each capsule learns a subspace and uses projection of input vector. & Better representation of information. Increased speed.\\
    RS-CapsNet & Splice convolutional feature map to create ResNet type architecture. & Allows stacking of capsule and increased non-linearity. \\
    SUPCAPS & Modified to use capsules as generator in unsupervised setting and added dropout. & Study on loss of capsule learning capabilities and importance of dropout for capsules.\\
    \bottomrule
  \end{tabular}
\end{table}

\section{Discussion}

    The focus on routing has led people to search for various types of routing procedures. The results on CIFAR10 \cite{zeiler2013stochastic} in \cite{hahn2019self} shows that EM routing performs worse than that of the dynamic routing. In contrast, EM routing on SmallNORB~\cite{lecun2004learning} gives similar results as that of Dynamic Routing, but the self-routing procedure beats them all.
    
    For KL Routing, experiments were done on a simple unsupervised perceptual grouping task~\cite{greff2017neural}, where KL Routing performs better than dynamic routing. There is no comparison between EM Routing and KL Routing, but given the type of task, it appears that EM routing will also perform well. Given the performance of EM routing on SmallNORB and Self Routing beating both Dynamic and EM Routing, Self Routing will likely perform well on such tasks. Furthermore, capsule networks give better performance when used in an ensemble~\cite{xi2017capsule}, supporting the Mixture-of-Experts approach towards routing.
    
    It has been shown that residual networks behave like ensembles of networks with different depths~\cite{veit2016residual} and the authors for Self Routing mentions integration with residual networks as future work. In contrast, the routing procedure of RS-CapsNet slices the convolutional feature maps to construct capsules, and the routing is done on these slices. This procedure can also be construed as doing Self Routing on complete input given to multiple capsules. Moreover, the slicing of input and then taking a linear combination of each is the same as projecting the complete input vector onto different vector subspaces. This approach is similar to CapProNet. Zhang et al.~\cite{zhang2018cappronet} apply the subspace capsules only as of the last layer with other networks like ResNet and DenseNet; their future work included the integration of the same in all layers. It is interesting to note that RS-CapsNet performs better in ensembles, so the same may hold for capsules networks with Self Routing.
    
    In EM routing, we previously mentioned that black box attacks gave performance similar to CNNs, which likely indicates that the methods used in white box methods~\cite{goodfellow2014explaining, kurakin2016adversarial} for generating adversarial examples are not effective against capsules or at least are not smart enough. The work is done by Michels et al.~\cite{michels2019vulnerability} shows that it is possible to fool capsule nets with sophisticated attacks, though they did modify the algorithms, and for simpler attacks where CNNs fail, capsule network does show robustness to attacks. Furthermore, their work only consider the capsule network proposed by Sabour et al.~\cite{sabour2017dynamic} but not the other variants, so it is not possible to generalize. Similarly, given the similarities between Self Routing and RS-CapsNet, it is possible to also be robust towards adversarial attacks, though this needs to be confirmed.
    
    Rawlinson et al.~\cite{rawlinson2018sparse} expects that sparse unsupervised capsules (SUPCAPS) will also show robustness similar to capsules with EM Routing because of the unsupervised training regime. Though this is just a speculation, it is based on the improved generalization when dropouts were added. It appears that the capsules need to be given a chance to explore other possibilities for learning the underlying hierarchical relationship given the same inputs, and it seems that dropouts are helping in this process.

    Autoencoder based capsule structure by Hinton et al.~\cite{hinton2011transforming} required to give information about the augmentation in image. In the work of Sabour et al.~\cite{sabour2017dynamic}, this problem appears to have been solved by using the margin loss and masking, but not specifically due to structuring of capsule. This is observable from the fact that when margin loss and mask is removed, the network starts to act as an autoencoder~\cite{rawlinson2018sparse}. The failure of stacked capsules~\cite{xi2017capsule} can be used to interpret that capsules in the hidden layer are not learning the latent information of features. Possibly, this is because capsules require margin loss to force them to explore different paths and thus capsules require the special supervision of a loss directly.

\section{Squash Function}
    The squash function plays the role of adding non-linearity in the routing process. An attempt to modify the squash function was done in ~\cite{xi2017capsule, wang2018optimization, huang2020capsnet}, but there has not been much work. In the previously mentioned routings and variants, equation \ref{eq:squash} is used or squashing itself is removed as in the case of EM Routing. We did not find a proper study of squash functions and hence did some experiments, though they were limited to not changing the direction of the vector. We propose the following variants of squash function:
    \begin{equation}
    \label{eq:squash_m}
    v_j = \frac{{||s_j||}_m^{m}}{1+{||s_j||}_m^{m}} \frac{s_j}{{||s_j||}_2}\,\,\,\, where \,\, m \in \mathbf{I}^+
    \end{equation}
    \begin{equation}
    \label{eq:squash_inf}
    v_j = \frac{{||s_j||}_\infty}{1+{||s_j||}_\infty} \frac{s_j}{{||s_j||}_2}
    \end{equation}
    
    For notation, we define the squash function represented by equation \ref{eq:squash_m} as $S_m$ and equation \ref{eq:squash_inf} as $S_\infty$. Our focus is on the relative performance of $S_1$, $S_2$, $S_3$, $S_4$, $S_5$, $S_{10}$ and $S_\infty$, for which we train the capsule network \cite{sabour2017dynamic} on MNIST \cite{lecun1998mnist} and CIFAR10 with five-fold validation.

\begin{figure}[!htp]
	\centering
	\includegraphics[width=1\linewidth]{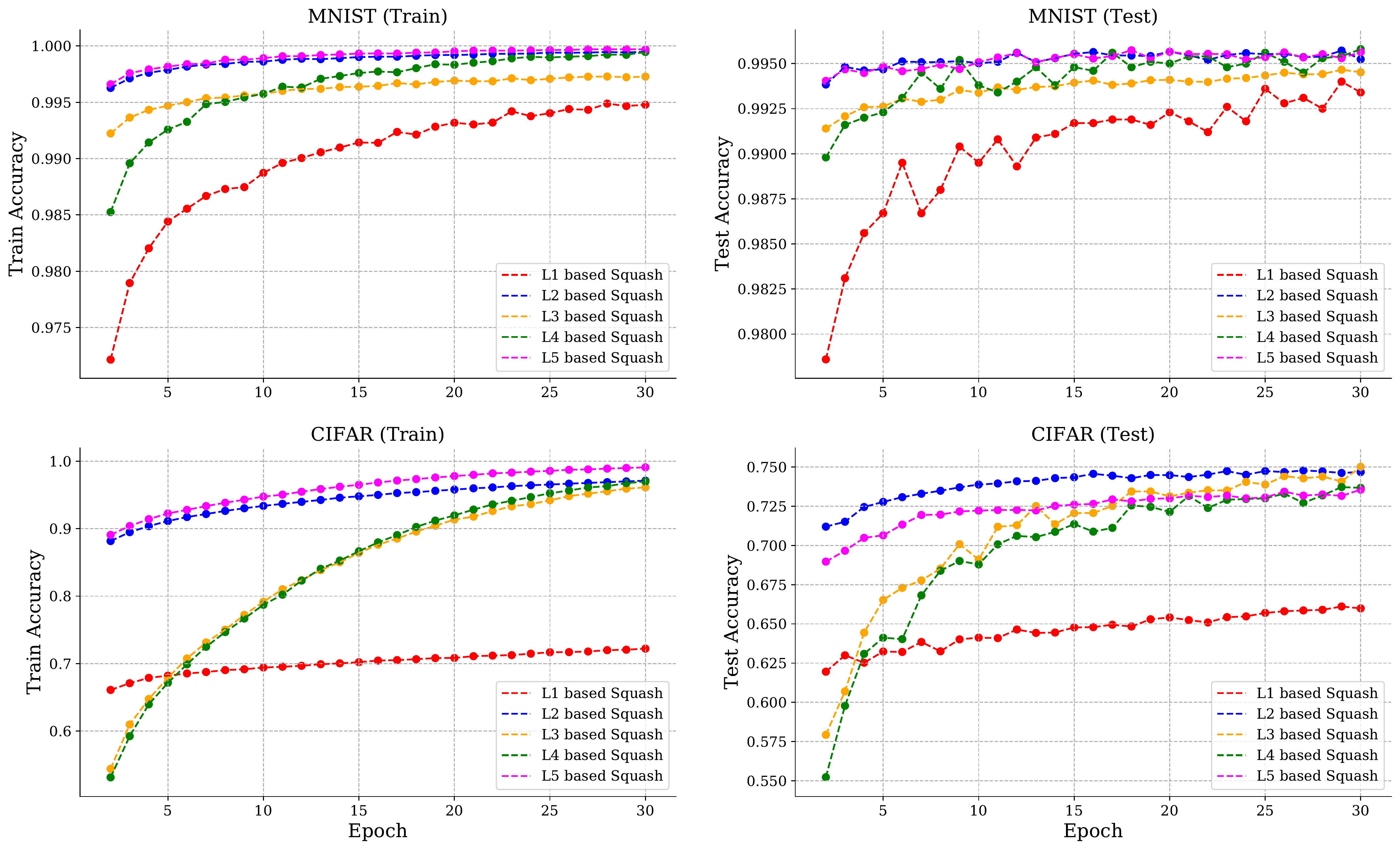}
	\caption{Performance comparison of different norm-based squash functions.}
	\label{fig:squash_loss}
\end{figure}





\begin{table}[!htp]
  \caption{ Results on MNIST and CIFAR10 }
  \label{squash_table}
  \centering
  \begin{tabular}{c  c c  c c }
    \toprule
    & \multicolumn{2}{c}{MNIST} & \multicolumn{2}{c}{CIFAR10} \\
    Function & Accuracy (\%) & Loss & Accuracy (\%) & Loss \\
    \midrule
    $S_1$ &  $99.408 \pm 0.095$ & $29.179\pm0.96$ & 66.108 $\pm$ 2.09 &  84.697 $\pm$ 1.48\\
    $S_2$ &  99.512 $\pm $ 0.028& 27.495 $\pm$ 0.49 & 74.525 $\pm$ 0.23& 74.038 $\pm$ 1.43\\
    $S_3$ &  99.417 $\pm $0.021 & 28.636 $\pm$ 0.58 & \textbf{75.068 $\pm$ 0.87}& \textbf{ 71.873 $\pm$ 1.59}\\
    $S_4$ &  \textbf{99.582 $\pm  $ 0.037}& \textbf{27.261$ \pm $0.31} & 73.727 $\pm$ 1.19 & 72.593 $\pm$ 1.62\\
    $S_5$ & 99.542 $\pm$ 0.037  &27.971 $\pm$ 0.38 & 73.506 $\pm$ 0.32& 77.401 $\pm$ 1.41\\
    $S_{10}$ &13.521 $\pm$ 0.021 & 98.473 $\pm$ 0.19  &  11.269 $\pm$ 0.27 & 121.713 $\pm$  0.85\\    
    $S_\infty$ &10.897 $\pm$ 0.013 & 114.513 $\pm$ 0.15  &  10.681 $\pm$ 0.21 & 132.612 $\pm$  0.79\\
    \bottomrule
  \end{tabular}
\end{table}

We observe that for $S_1$, the train accuracy remain close to that of test but the overall performance is worse than $S_2$. In contrast, for $m > 2$, the train accuracy starts to shoot while the test accuracy is close to $S_2$. The likely reason for this is that with increasing $m$, votes with little increase in magnitude can give very high agreement. However, when the magnitude drops below unity, the vote is highly penalized. As $m \rightarrow \infty$, the final magnitude as the function of original magnitude will have a jump from $0$ to $1$ at value unity.

For $S_\infty$ we find that the capsules miserably fail to generalize. This is likely because $S_\infty$ fails to penalize disagreements as well as to give high weight to agreements. The same logic can be extended towards the family $S_m$. But this logic is limited in the sense that if we highly penalize disagreements, then once a capsule has decided a route in training, it will fail to discover others.

We observe that for $S_2$ and $S_5$, the training accuracy shoots up in the initial epochs itself and the similar can be seen in test accuracy also, but this is not the case for $S_3$ and $S_4$. This shows that the value of $m$ is deciding how much previously learned knowledge the network will preserve, but no clear relation is visible. For $S_2$, we expect that the network will moderately penalize disagreements while the case of $S_5$ is the other extreme. The behaviour of $S_3$ and $S_4$ is somewhat in the middle and the slow learning is likely because the network does not easily form or forget an opinion in the initial stages.

\section{Conclusions}
    
    It is evident that capsules require extreme supervision for specializing them, thus limiting the possibility of achieving better performance from deep capsule networks. Dropouts, residuals and subspaces provide a way to solve this problem. We observe that there needs to be proper comparison between various routing methods on multiple types of datasets as certain routing methods work good only for specific tasks. Also, there is little work on the squash function which primarily brings non-linearity in capsules. Our work on squash function shows how greatly it can affect the overall learning process of capsule networks, thus, an analysis on finer components of these networks are worth an investigation.
    
    
\section{Future Work}
    
    We believe that the on key problem in capsule networks regarding the specializing comes up because the training of the complete network is done in an end-to-end manner. In contrast, it is possible that capsules are added slowly to a capsule layer. Further, we think that incorporating techniques similar to Rawlinson et. al.~\cite{rawlinson2018sparse}, it is possible to force the network to explore other routes for learning the underlying hierarchical relationships with the newly added capsules. Moreover, using subspaces can allow forcing capsules to learn specific features, though it might be necessary to add constraints on learning of subspaces to ensure that all capsules learn a different one. Also, incorporating routing methods like Self Routing can mitigate the issue of previous training that did not include the vote of the newly added capsule. This can be corresponded with the concept of incremental learning that is done in CNNs \cite{roy2020tree}. Incorporating techniques like incremental learning in capsule networks is just like putting clay slowly to add new pathways to the previously learned representations between capsules. 

\newpage
\bibliographystyle{plain}
\bibliography{ref}
\medskip     

\end{document}